\documentclass[letterpaper, 10 pt, conference]{ieeeconf}  

\IEEEoverridecommandlockouts                              

\usepackage{times}
\usepackage{epsfig}
\usepackage{graphicx}
\usepackage{amsmath}
\usepackage{amssymb}
\usepackage{caption}
\usepackage{subcaption}
\usepackage[breaklinks=true,bookmarks=false, colorlinks=true]{hyperref}

\title{\LARGE \bf
Polarimetric Monocular Dense Mapping \\Using Relative Deep Depth Prior*
}

\author{Moein Shakeri$^{1}$, Shing Yang Loo$^{1}$, Hong Zhang$^{1}$
\thanks{*This work was supported in part by UAHJIC.}
\thanks{$^{1}$Moein Shakeri, Shing Yan Loo and Hong Zhang are with the Department of Computing Science, University of Alberta, Edmonton, Canada {\tt\small shakeri@ualberta.ca}; {\tt\small lsyan@ualberta.ca}; {\tt\small hzhang@ualberta.ca}}
}

\begin{document}

\maketitle
\thispagestyle{empty}
\pagestyle{empty}

\begin{abstract}

This paper is concerned with polarimetric dense map reconstruction based on a polarization camera with the help of relative depth information as a prior. In general, polarization imaging is able to reveal information about surface normal such as azimuth and zenith angles, which can support the development of solutions to the problem of dense reconstruction, especially in texture-poor regions. However, polarimetric shape cues are ambiguous due to two types of polarized reflection (specular/diffuse). Although methods have been proposed to address this issue, they either are offline and therefore not practical in robotics applications, or use incomplete polarimetric cues, leading to sub-optimal performance. In this paper, we propose an online reconstruction method that uses full polarimetric cues available from the polarization camera. With our online method, we can propagate sparse depth values both along and perpendicular to iso-depth contours. Through comprehensive experiments on challenging image sequences, we demonstrate that our method is able to significantly improve the accuracy of the depthmap as well as increase its density, specially in regions of poor texture.

\end{abstract}

\vspace{-3pt}
\section{INTRODUCTION}

Incremental 3D dense map reconstruction from an image sequence is a fundamental problem with implication in a variety of computer vision and robotics applications, such as object recognition~\cite{intro_2,intro_3} and navigation~\cite{intro_1}, and can improve the overall performance of the traditional methods in challenging situations~\cite{shakeri_ICCV, shakeri_CVPR, shakeri_CCC} due to use of depth features. The problem has been extensively studied and many solutions proposed in the last decade, especially in object detection and visual SLAM (simultaneous localization and mapping) research.   
However, almost all existing solutions have some significant drawbacks including unreliable reconstruction in texture-poor regions, reliance on direct depth sensing (e.g., RGB-D) or generalization (e.g., in learning based methods).  

On the other hand, optimization based polarimetric methods are well-known to address these drawbacks, and they have been widely studied~\cite{wu2020hdr,polar4,polar7,polar8}. Polarization is a property of light, which can convey rich geometric information about the environment. However, due to the difficulty of capturing useful polarization images by a moving camera, existing polarimetric methods have been impractical in such applications as robot navigation. 
Recent advances in imaging technology has led to the introduction of polarization cameras that employ multi-directional on-chip linear polarizers, to enable the use of polarimetric methods in robotics on a moving platform. Nonetheless, existing polarimetric methods need strong assumptions about the environment and the objects, and failure to meet those assumptions causes uncertainties and ambiguities on the obtained polarization information. It is these ambiguities that make the dense reconstruction from polarization cues alone challenging.
\begin{figure}
\includegraphics[width=\linewidth]{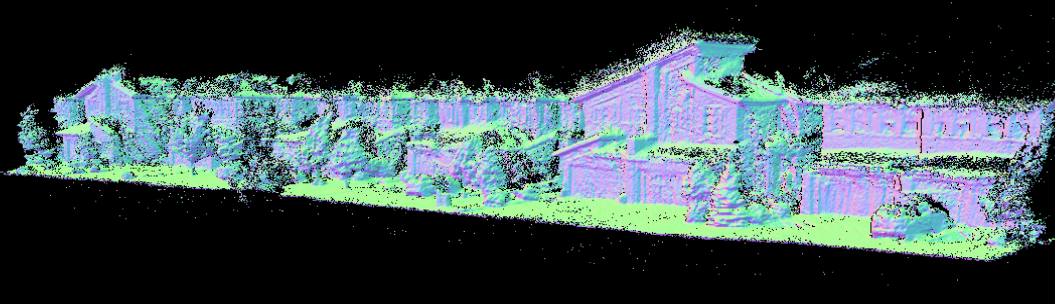}
\caption{Sample resonstruction result of the proposed method. Surface normal map is shown in pseudo-color.}
\label{sample_3d}
\vspace{-10pt}
\end{figure}

In this paper, we propose an incremental solution to the problem of dense mapping using a polarization camera. 
Specifically, we first initialize a rough depthmap with a sparse visual SLAM method and extract points with reliable depth values. We then use a relative depthmap estimated by a neural network to resolve the ambiguities of the polarimetric information obtained from the polarization camera. Finally we use a novel iterative process to a) propagate the valid sparse depth values along a proper direction using the polarimetric cues, b) estimate the depth variation within the depth gradient map, and c) smooth the propagated/estimated depth values in the featureless regions to improve map accuracy and increase its density. 
We have evaluated the proposed method on both indoor and outdoor scenes captured by a polarization camera. One sample result of our proposed method is shown in  Fig.~\ref{sample_3d}. 
The main contributions of this paper are as follows.
\begin{itemize}
\item{We propose to use a relative depthmap to resolve the ambiguities of the polarimetric information. Due to the dense relative depthmap, this approach is able to disambiguate the polarimetric data of all the pixels.}
\item{We introduce an iterative depth propagation and smoothing process, which uses full polarimetric information with the help of relative depth prior versus partial polarimetric information. This iterative process reconstructs the dense map with more accurate points in comparison with the state-of-the-arts.}
\item{Due to the lack of polarization image sequences for SLAM, we create synthetic polarization image sequences from ICL-NUIM depthmaps. We simulate polarization images using the Blinn-Phong reflectance model under an arbitrary point light source, and create four polarization image for each regular depthmap. To our knowledge, this is the first synthetic polarization image dataset with varying albedo for SLAM.}
\end{itemize}

The remainder of this paper is organized as follows. Related works on dense mapping and the theory of polarization are summarized in Section~\ref{related_work}. Section~\ref{proposed_method} explains the details of our polarimetric monocular dense reconstruction method. Experimental results and discussion are presented in Section~\ref{experiment}, and concluding remarks in Section~\ref{conclusion}.
\section{Related Works}
\label{related_work}
In this section we first review related works on dense reconstruction and then explain relevant polarization theory that is the basis of our proposed method. 
\subsection{Dense Reconstruction}
Monocular dense reconstruction from a scequence of images is a well-studied area of research and many solutions have been proposed for various applications in computer graphics, computer vision and robotics. One traditional and major group of these solutions is ``structure-from-motion" (SfM), which has seen tremendous progress over the years~\cite{Related1_SFM,Related2_SFM, Related3_SFM,Related4_SFM,Related5_SFM,Related6_SFM,colmap}. While the existing methods have demonstrated their feasibility, robustness and completeness (i.e., density) remain key challenges in incremental SfM methods to prevent their practical use especially for reconstructing environments with texture-poor regions. 

The second group of methods relies on specialized imaging technologies such as RGB-D cameras~\cite{kinectfusion} or polarization cameras~\cite{polarized_camera} to build a dense map. Methods with RGB-D cameras are scalable and work in real-time~\cite{kinectfusion,RGBD1,RGBD2}; however, they limit the type of environments to short-range indoor scenes. Methods with polarization cues have been exploited in many 3D reconstruction algorithms~\cite{polar1, polar4, polar5, polar6,polar9, polar10}. Early methods~\cite{polar7,polar8} use geometric priors (e.g., the surface normal on the boundary and convexity of the objects) for shape reconstruction. Polarimetric information has also been combined with shape-from-shading (SfS) to solve the ambiguity in surface normal estimation in order to recover 3D shape~\cite{polar5}. Methods based on polarization cues assume that the incident illumination is unpolarized and so they can be even used for outdoor scenes. 
Recently, some studies demonstrated the use of polarimetric information for camera localization~\cite{polar3} and incremental dense mapping~\cite{polar2}. Yang {\it{et al.}}~\cite{polar2} proposed to solve monocular SLAM by incorporating photometric and polarimetric information to recover the surfaces of indoor objects in featureless regions with promising results. Their method iteratively propagates valid depth values along iso-depth contours, and relies on the PatchMatch~\cite{gipuma} algorithm to estimate the map around the propagated points. However, there is no constraint on the depth variation between iso-depth contours, and this may cause the loss of the depth consistency along the depth gradient even with the help of smoothing optimization, specially in large textureless regions. We will discuss this issue further in Section~\ref{iterative_process}.

The third group of methods takes advantage of the power of convolutional neural networks to directly regress scene depth from an input image. Several architectural innovations have been proposed to enhance prediction accuracy~\cite{deep1,deep2,deep3,deep4,deep5,deep6}. These methods tend to work well in scenes similar to those in which the neural network is trained, but do not generalize well to dissimilar scenes, due to the limited scale and diversity of the training data. To address the generalization issue, relative depth prediction methods trained on combined datasets of diverse scenes have been proposed~\cite{deep7,MiDaS}. A clear drawback is that these methods fail to recover the exact geometric 3D shape as only ordinal relations are predicted. Among the leading methods for relative depth prediction ``in the wild", one effective solution is recently proposed by Ranftl {\it{et al.}}~\cite{MiDaS}. This method relies on datasets with diverse environments and on novel loss functions that are invariant to the major sources of incompatibility between datasets, including unknown and inconsistent scale and baselines. Although this predicted relative depth is essentially constrained to the disparity space and cannot recover the exact geometric shape, the obtained results show consistent relative depth values for each individual surface as shown in Fig.~\ref{fig:relative_depth}. We propose to use this depth consistency as a prior to resolve the ambiguities of polarimetric information as will be explained in Section~\ref{disambiguation}. 
\vspace{-3pt}
\subsection{Polarization Theory}
\label{polarization_info}
\vspace{-2pt}
Our proposed method is based on a polarization camera, which implements pixel-level polarization filters and has resulted in real-time and high resolution measurment of incident polarization information. Each calculation unit of the camera consists of four pixels and uses four on-chip directional polarizers, at 0, 45, 90, and 135 degrees, to capture four perfectly aligned and polarized images. The images can be analyzed to estimate the degree of linear polarization (DoLP) and angle of linear polarization (AoLP) of the irradiance received at the camera lens. The degree of polarization is defined as follows.
\begin{equation}
\label{dop}
\vspace{-2pt}
\rho=\frac{I_{max}-I_{min}}{I_{max}+I_{min}}
\end{equation}
where $I_{max}$ and $I_{min}$ are the maximum and the minimum measured radiance at every pixel. These two parameters and the angle of polarization $\phi$ are unknown and can be computed using three or more polarization images with different but known polarizer filter angles. Since our camera can capture four images at once, the computation of DoLP and AoLP is straightforward and has in fact a closed form solution~\cite{polar12}.

DoLP and AoLP provide constraints on the surface normal of each space point. These constraints depend on the polarization model, which can characterize either diffuse reflection or specular reflection. In this paper we follow recent works~\cite{polar6, polar3, polar2} and assume that reflection from scene points at each pixel can be classified as either diffuse dominant or specular dominant. 

\subsubsection{Azimuth angle estimation}
\label{azi_est}
The azimuth angle $\varphi$ of the surface normal represents the angle between the projected surface normal direction and $x-$axis of the 2D image plane. For diffuse dominat reflectance, AoLP $\phi$ determines the azimuth angle $\varphi$ up to a $\pi$ ambiguity~\cite{polar5}, as follows.
\begin{equation}
\label{eq:diffuse_azi}
\vspace{-2pt}
\varphi=\phi \,\,\,or\,\,\, \varphi=\phi+\pi 
\vspace{-2pt}
\end{equation}
and, for specular dominant reflectance, the azimuth angle is given by:
\begin{equation}
\label{eq:spec_azi}
\vspace{-5pt}
\varphi=\phi\pm\pi/2
\end{equation}

These relations also introduce $\pi/2$-ambiguity due to two different reflection models (i.e., there is a $\pi/2$ difference between~(\ref{eq:diffuse_azi}) and~(\ref{eq:spec_azi})).   
\subsubsection{Zenith angle estimation}
\label{zenith_est}
The zenith angle $\theta$ of the surface normal represents the angle between the surface normal and the viewing direction from the camera, and it is related to DoLP. For diffuse reflection, DoLP $\rho$ is related to the zenith angle $\theta\in[0,\pi/2]$ as follows~\cite{polar5}.
\begin{equation}
\label{eq:diffuse_zenith}
\vspace{-5pt}
\hspace{-6pt}\rho=\frac{(\eta-\frac{1}{\eta})^2sin^2\theta }{4cos(\theta)\sqrt{\eta^2-sin^2\theta}-(\eta+\frac{1}{\eta})^2sin^2\theta + 2\eta^2+2}
\end{equation}
where $\eta$ is the refractive index of the object surface. Since $\rho$ is defined in~(\ref{dop}), with a known $\eta$, zenith angle $\phi$ for pixels with diffuse reflection has a closed-form solution.

For specular reflection, the relationship between $\theta$ and $\rho$ is given by:
\begin{equation}
\label{eq:spec_zenith}
\rho=\frac{2sin^2(\theta)cos(\theta)\sqrt{(\eta^2-sin^2\theta)}}{\eta^2-sin^2\theta-\eta^2sin^2\theta+2sin^4\theta}
\end{equation}
From~(\ref{eq:spec_zenith}), there are two solutions to $\theta$, which normally cause an additional ambiguity in the specular case. However, in our method, we are able to resolve this ambiguity by using a relative depth prior to choose the correct solution.
\section{Proposed Method}
\label{proposed_method}
\begin{figure}
\vspace{3pt}
\includegraphics[width=\linewidth, height=3.7cm]{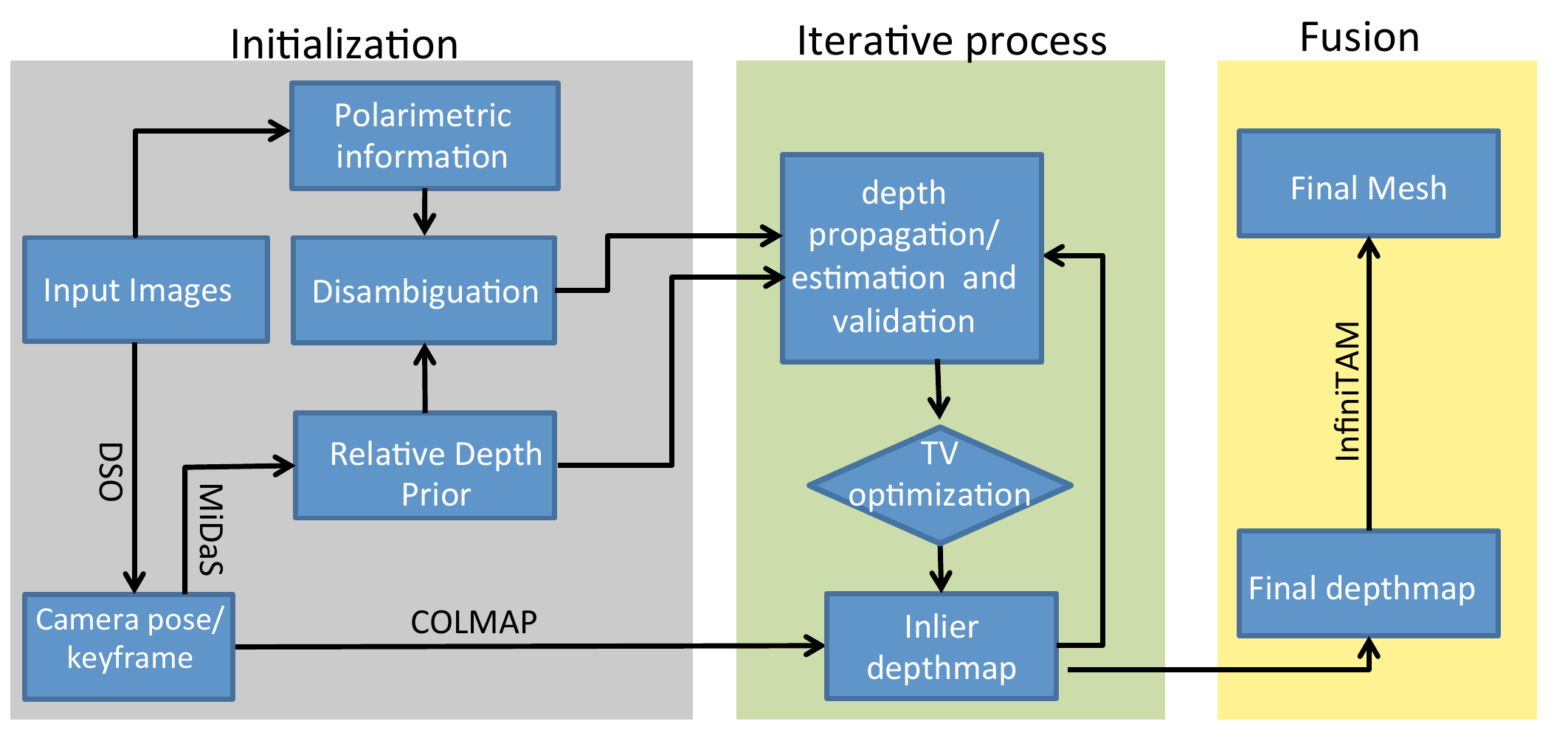}
\caption{Overview of our proposed polarimetric method}
\vspace{-10pt}
\label{fig:scheme}
\end{figure}
 Fig.~\ref{fig:scheme} shows the overview of our proposed reconstruction method. It takes as input a sequence of images captured by a moving polarization camera. We use a visual odometery (VO) method to define the keyframes and estimate the corresponding camera poses, which are essential for our map reconstruction method. Then we initialize a depthmap per keyframe using the COLMAP algorithm~\cite{colmap}. The depth consistency check between keyframes produces a set of inlier 3D points; however, those points are sparse and available mostly in the textured regions. Depth estimation of textureless regions is a challenge in dense reconstruction as was discussed in Section~\ref{related_work}. To address this issue, we propose to use relative depth as a prior~\cite{MiDaS} per keyframe to provide a consistent depth gradient for each individual surface in a scene. Based on this prior, we are able to resolve the ambiguities of polarimetric cues. Then we use the disambiguated polarimetric cues and the depth prior to densify the sparse depth map in an iterative process. Finally we fuse the dense depthmaps incrementally over multiple keyframes to reconstruct the final map.
 
\subsection{Depthmap initialization}
\label{initialization}
In our polarimetric reconstruction method, the first step is camera pose estimation to localize the camera. We use the direct sparse odometry (DSO) algorithm~\cite{DSO}, a state-of-the-art VO method, to generate camera poses and keyframes sequentially. At each camera pose, we take the mean of the four polarized channel intensities to generate a grayscale image as input to DSO. Then we use COLMAP~\cite{colmap} to reconstruct an initial depthmap. Although we could manipulate DSO to get an initial sparse depthmap, the map would be too sparse in comparison with the map from COLMAP specially in texture-rich regions with complex geometry, which hinders the propagation of depth values due to the change in azimuth angle. In contrast, since COLMAP incorporates both photometric and geometric constraints in a joint optimization, the obtained depthmap is more complete and more accurate in comparison with that from DSO or the competitor~\cite{gipuma} used in~\cite{polar2}. 

As notations that will be used later, let $K_t$ be the current keyframe. Then we use three neighboring keyframes $K_t, K_{t-1},$ and $K_{t-2}$ and their camera poses to compute the initial depthmap $z_t$ for $K_t$. The COLMAP algorithm can be efficiently executed on a GPU to initialize the depthmap in real-time.

\begin{figure}
  \centering
  \vspace{2pt}
  \includegraphics[width=0.7\linewidth, height=2.5cm]{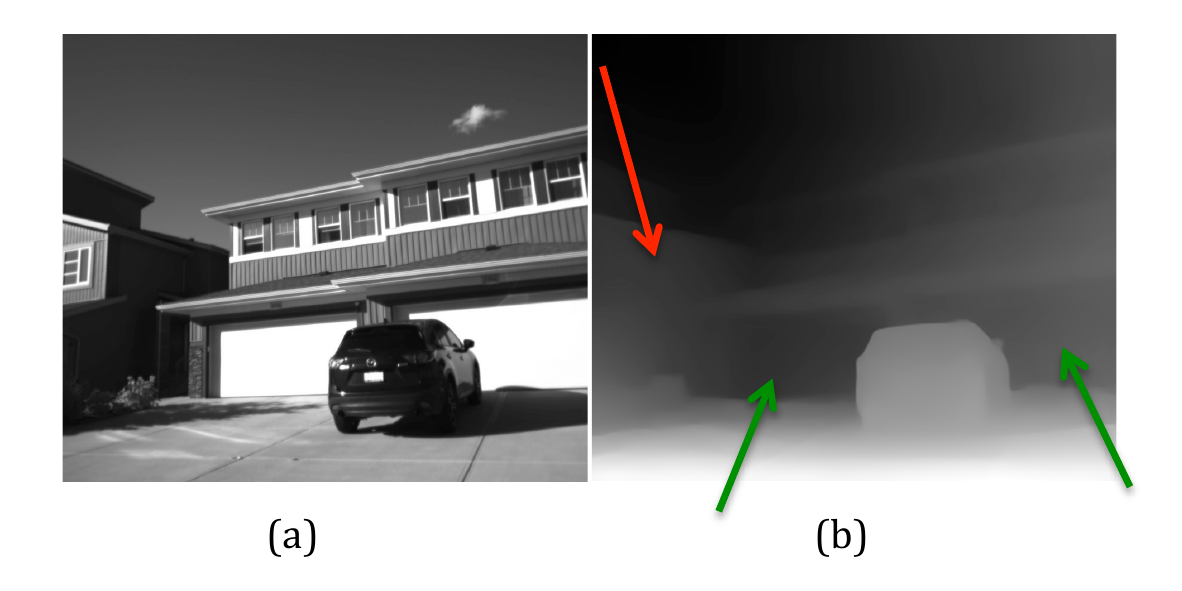}
  \vspace{-5pt}  
  \caption{(a) Original input image and (b) estimated relative depth using~\cite{MiDaS} }
  \vspace{-15pt}
  \label{fig:relative_depth}
\end{figure}

In parallel, we use MiDaS~\cite{MiDaS}
on the current keyframe to compute a relative depthmap $z'_t$. As discussed in Section~\ref{related_work}, this relative depthmap is expressed in the disparity space with an unknown depth scale, and this makes map reconstruction from a relative depth estimate alone a highly non-linear problem. In addition, sometimes the predicted ordinal relations between surfaces with different directions may be incorrect, as shown in Fig.~\ref{fig:relative_depth} where the red arrow shows a region which has a wrong predicted relative depth in contrast to the regions defined by the green arrows. Specifically, unlike the predicted relative depth, garage doors in this example should be much closer to the camera than the side wall. However, the predicted relative depth still can be helpful since it is smooth and consistent within each local surface. We use this property of $z'_t$ to solve the ambiguities of polarimetric information, as will be explained in the next section.
\subsection{Disambiguation of polarimetric cues}
\label{disambiguation}
Azimuth angle at a point on a surface provides a strong constraint on the geometry of the surface, and enables us to propagate depth in the appropriate direction. As discussed in Section~\ref{polarization_info}, polarized images can determine the azimuth angle ($\varphi$) of the surface normal with exactly two types of ambiguity: the $\pi$-ambiguity and the $\pi/2$-ambiguity.~\cite{polar1} showed that, for each iso-depth contour, the azimuth angle of each point is perpendicular to that contour, and therefore the $\pi$-ambiguity is not an issue for depth propagation. 


For resolving $\pi/2$-ambiguity, Cui {\it{et al.}}~\cite{polar1} proposed a graph-based optimization solution, which is computationally too expensive for real time applications. Yang {\it{et al.}}~\cite{polar2} proposed a faster method for azimuth angle disambiguation by tracing iso-depth contours for the two possible azimuth angles, and picking the contour with the smaller variance in depth values based on the initial depthmap. However this process fails where the initial depth values are missing. 

In our method, we use the obtained relative depthmap $z'_t$ for resolving $\pi/2$-ambiguity of azimuth angles. Let $n'(p)$ be the estimated surface normal at pixel $p$, formulated via surface gradient as follows.
\begin{equation}
\label{eq:normal_depth}
n'(p) = 
\begin{bmatrix}
-f\times\nabla_xz'_p \\ -f \times\nabla_yz'_p \\ (x_p-x_0)\nabla_xz'_p+(y_p-y_0)\nabla_yz'_p+z'_p
\end{bmatrix}
\end{equation}
where $p=(x_p,y_p)$. $(x_0,y_0)$ and $f$ are the principal point and the focal length of the camera, respectively. We denote by $\bar{n}(p)=\frac{n'(p)}{\|n'(p)\|}$ the unit surface normal, which can be expressed in the camera coordinate system as follows.
\begin{equation}
\label{eq:normal_polarize}
\bar{n}(p) = \frac{n'(p)}{\|n'(p)\|}=
\begin{bmatrix}
cos\varphi sin\theta \\ sin\varphi sin\theta \\ cos\theta
\end{bmatrix}
\end{equation}
 
From~(\ref{eq:normal_depth}) and~(\ref{eq:normal_polarize}), the depthmap gradient should be consistent with respect to the azimuth angle, i.e. $tan(\varphi_p)=\nabla_yz'_p/\nabla_xz'_p$. Therefore, we can recover the proper azimuth angle by considering the following alignment error. 
\begin{equation}
\label{eq:azi_disambiguation}
\|\nabla_yz'_p/\nabla_xz'_p-tan(\varphi_p)\|^2
\end{equation}
In~(\ref{eq:azi_disambiguation}) we only need to check four available choices and select the one that minimizes the alignment residual. This approach is fast and robust to depth errors since we only use it to select the best angle from the four candidates. Besides, this process can also handle $\pi$-ambiguity of azimuth angle.

Solving~(\ref{eq:azi_disambiguation}) also labels the reflectance type at each pixel as specular or diffuse, and this helps us to use a proper model for zenith angle estimation. From~(\ref{eq:diffuse_zenith}) and~(\ref{eq:spec_zenith}), zenith angle estimation needs known refractive index $\eta$. Fortunately the dependency on $\eta$ is weak~\cite{polar5} and its value for dielectric objects ranges between 1.4 and 1.6. We assume $\eta=1.5$ for the rest of this paper. For pixels with dominant specular reflection, we still need to minimize $\|\theta(p)-\theta'(p)\|$ to
select one of the two answers that satisfies~(\ref{eq:spec_zenith}). 

Now, in order to use the polarimetric cues for depth reconstruction, we can propagate depths from a set of inlier sparse map points from COLMAP to featureless regions with unknown depth values.

\subsection{Inlier point extraction}
\label{validation}
The overall accuracy of depth propagation depends on an accurate initialization of the sparse points. In our method we first perform a two-view consistency check between keyframes $K_t$ and $K_{t-1}$ similarly to~\cite{polar2}. Particularly, we reproject $z_{t-1}$ (the depthmap computed at the previous keyframe) into the current keyframe $K_t$ and filter out depth values where the depth difference is more than a threshold, defined as $0.01 \times (z_{max}-z_{min})$ in our method where $[z_{max}, z_{min}]$ is a predefined depth range~\cite{colmap}. In the next section, we will explain the process of depth propagation/estimation using the obtained inlier points $z_t$.

\subsection{Iterative depth propagation, estimation and smoothing}
\label{iterative_process}
Previous methods have shown that depth values should be constant along a direction perpendicular to the azimuth angle~\cite{polar1,polar2,polar11}. Therefore, depth propagation can be carried out by tracing the reliable sparse points along the two directions perpendicular to the azimuth angle. 

Let $\varphi_p$ be the azimuth angle of pixel $p$ with known depth value. We propagate the depth of $p$ in the two directions:
$\overrightarrow{d_+}: [cos(\varphi_p + \pi/2), sin(\varphi_p + \pi/2)]$ 
and $\overrightarrow{d_-}: [cos(\varphi_p - \pi/2), sin(\varphi_p - \pi/2)]$.
The propagation process stops once the change in the azimuth angle between two neighboring pixels is larger than a threshold ($\pi/6$ in our experiments) to avoid propagation at a depth discontinuity. However, in scenes with large featureless regions, error in azimuth angle can drift significantly after a couple of iterations in depth propagation. This error is more pronounced in real applications, since real-world objects usually have mixed reflection (i.e., a combination of diffuse and specular reflection). This can be even more crucial in incremental methods due to the lack of multi-view optimization such as bundle adjustment over a window images.  

To address this issue and to reduce the error in our method, we also estimate depth values along the azimuth angle which is consistent with the depthmap gradient slope (maximum depth variation). As explained in Section~\ref{initialization}, depth gradient from the relative depth prior is smooth and consistent for each individual surface. We use this consistency to provide a constraint on the slope of the propagated depth values in different regions of one surface. 
From the first two rows of~(\ref{eq:normal_depth}) and~(\ref{eq:normal_polarize}), depth gradient is related to the zenith angle by $(\nabla^2_xz+\nabla^2_yz=\frac{sin^2\theta \times \|\bar{n}\|}{f^2})$. Therefore, we first adjust the relative depth gradient in the two directions of gradient vector (which is along the azimuth angle) based on the zenith angle $\theta$ at pixel $p$. Then we compute the actual depth gradient and estimate the depth values up to a scale using inlier depth at $p$. Both depth gradient adjustment and depth estimation are combined using the following equations. 
\vspace{-3pt}
\begin{equation}
\label{eq:zenith_adjustment}
\begin{split}
z_{p^+}=\frac{z'_p+\frac{sin\theta_p}{sin\theta'_p}\times \triangle_+ z'_p}{z'_p}\times z_p \\
z_{p^-}=\frac{z'_p+\frac{sin\theta_p}{sin\theta'_p}\times \triangle_- z'_p}{z'_p}\times z_p \\
\end{split}
\end{equation}
where $\theta'$ is the zenith angle computed via relative depthmap $z'$ as follows.
\vspace{-2pt}
\begin{equation}
\vspace{-1pt}
\label{eq:zenith_rel}
\theta'(p)= arccos(\bar{n}(p). v(p))
\end{equation}
In~(\ref{eq:zenith_rel}), $v(p)=-[\frac{x_p-x_0}{f}, \frac{y_p-y_0}{f}, 1 ] / \|\frac{x_p-x_0}{f}, \frac{y_p-y_0}{f}, 1]\|$ is the vector pointing towards the viewer from a point on the surface. 
$\triangle_- z'_p$ and $\triangle_+ z'_p$ are depth differences between $z'_p$ and its neighbors ($p^+$ and $p^-$) along azimuth angle in two directions.~(\ref{eq:zenith_adjustment}) estimates the depth of neighboring pixels of $p$ with the consistent scale. This depth estimation step keeps the gradient consistency between different propagated regions of one surface.


After the propagation and the estimation steps, we validate the new depth values using a similar approach as Section~\ref{validation}. However, for scenes with large featureless regions, there may still be pixels with unreliable depth after the validation process.
Since featureless regions should have a smooth depth variation, we use the following approach to further optimize the depth for all pixels with known depth.

\begin{equation}
\label{eq:smoothing}
min_z \frac{1}{2}\|z-z^t\|^2+ \lambda J(z)
\end{equation}
where $J(z) = \sum_p |\tau_p\nabla z_p|$ and $\lambda=0.3$ is a regularization parameter that controls the smoothness.  We adopted $\tau_p=e^{-\zeta |\nabla I_p |}$ from~\cite{polar2} where $\nabla I_p$ is the image gradient at $p$ and $\zeta = 3$ in our experiments.~(\ref{eq:smoothing}) is in a standard form of total variation minimization and can be solved efficiently by~\cite{TV_opt}.
The process of depth propagation/estimation, validation and smoothing iterates until convergence, i.e., when the ratio between the number of newly added reconstructed points and the total number of reconstructed points at the current keyframe is less than a threshold (0.1 in our experiments). These iterative steps provide a relatively dense depthmap for each keyframe.
Finally, we use the InfiniTAM fusion algorithm~\cite{infinitam} to combined the depthmaps $z$ from~(\ref{eq:smoothing}).


\section{Experimental results and discussion}
\label{experiment}
In this section we present the experimental results of our proposed method on both synthetic and real polarization image sequences. Specifically we evaluate our method in two ways. In the first set of experiments, we evaluate our method quantitatively in terms of map density and accuracy by comparing the results with those from three visual SLAM methods DSO~\cite{DSO}, SVO~\cite{SVO}, and LSD-SLAM~\cite{lsd-slam} on synthetic and real polarization image sequences. In the second set of experiments we show the qualitative map reconstruction results of the proposed method. To show the capability of the proposed method, we also compare the reconstructed results of our method with deep-learning based algorithms on both indoor and outdoor polarization image sequences qualitatively.

\subsection{Evaluation of map density and accuracy}
\begin{figure}[!t]
  \centering
  \vspace{5pt}
  \includegraphics[width=0.9\linewidth, height = 4.5cm]{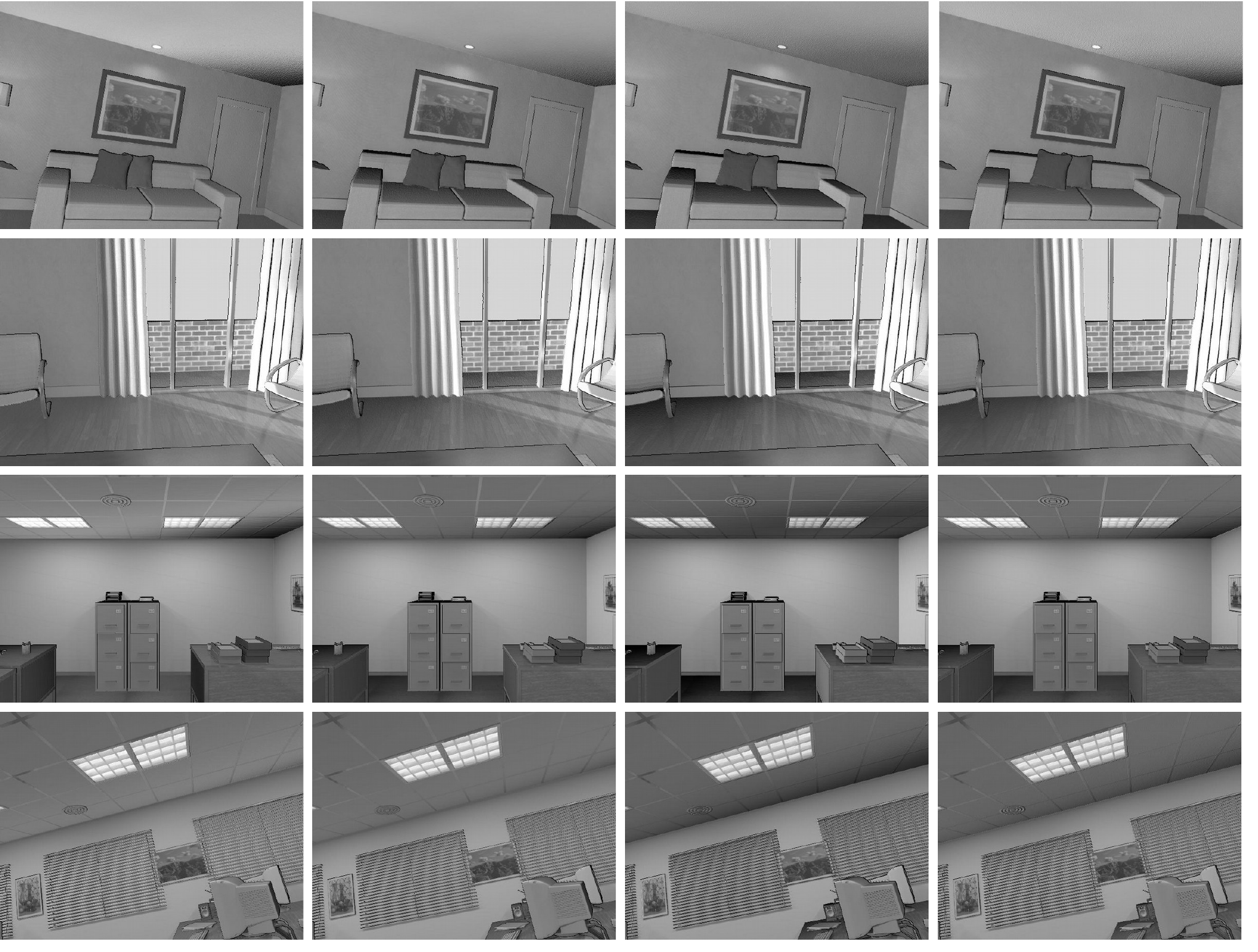}  
  \caption{Synthetic polarization images. Each row: four created polarized channel intensities of one synthetic image. Top to bottom: created synthetic polarization images using ``living room-kt0", ``living room-kt1", ``office room-kt1", and ``office room-kt3" sequences in ICL-NUIM dataset.}
  \label{fig:synthetic_pol_data}
  \vspace{-1em}
\end{figure}

To overcome the lack of polarization image datasets for scene reconstruction, we have had to first create our own synthetic polarization image sequences, and then evaluate our proposed method on them. To create our synthetic dataset, we use four sequences of a popular SLAM dataset, the ICL-NUIM dataset~\cite{synthetic_dataset}. We render the depth images with the Blinn-Phong reflectance model under an arbitrary point lighting source $s$ using the pinhole camera model, using the intensity values of synthetic images as albedo texture. Based on Section~\ref{polarization_info} and~\cite{polar5}, sample polarization images with polarizer filter angles at $0^{\circ}, 45^{\circ}, 90^{\circ},$ and $135^{\circ}$ are shown in Fig.~\ref{fig:synthetic_pol_data}. We then apply our method to this dataset and compare the reconstruction results with those obtained from the competing visual SLAM methods. Table~\ref{synthetic_evaluation} shows the results of the comparison, using the mean absolute relative error ($AbsRel = 1/M \sum_M{\|z-z_{gt}\|/z_{gt}}$) and the average number of reconstructed map points per image as the performance metrics. Clearly, the proposed method is able to outperform the competing methods in terms of the accuracy and density of the reconstructed points. 

\begin{table}[b]
\vspace{-5pt}
\caption{Comparison of the absolute relative error and the average number of reconstructed points on all sequences between the proposed method and the competing methods.}
\resizebox{\linewidth}{!}
{\begin{tabular}{l c c c r}
\hline
Methods         & LSD-SLAM & SVO & DSO & Ours \\
\hline
Living room kt0 & 0.1573 & ------ & 0.0864 & \textbf{0.0713} \\
Living room kt1 & 0.1447 & 0.1299 & 0.0868 & \textbf{0.0602} \\
Office room kt1 & 0.1274 & ------ & 0.0791 & \textbf{0.0720} \\
Office room kt3 & 0.2108 & 0.1622 & \textbf{0.0776} & 0.0793 \\
\hline
Ave. $\#$ of points per image  & 96673  & 284 & 1402 & \textbf{162784} \\
\hline
\end{tabular}}
\label{synthetic_evaluation}
\end{table}

We also evaluate the effect of the main steps of our proposed method using the same two metrics as above on the ``Living room kt1" polarization sequence. Table~\ref{ablation_evaluation} illustrates the results of this experiment. The second column in the table shows the accuracy of inlier points after validating the initialized points by COLMAP (Section~\ref{validation}). This step shows significant increase in the number of reconstructed points with less error in comparison with the DSO baseline; however, the COLMAP map remains sparse. These inlier points are then used in the iterative propagation/smoothing process to densify the map. After the first and the seventh iterations of the propagation and smoothing process (Section~\ref{iterative_process}), the number of reconstructed map points grows by over an order of magnitude with a slightly increased but acceptable reconstruction error. Compared with DSO, the proposed method is far superior in terms of both the accuracy and the number of the reconstructed map points.
\begin{table}[t]
\vspace{5pt}
\caption{Effect of different steps in our algorithm on Living room kt1 polarization sequence}
\resizebox{\linewidth}{!}
{\begin{tabular}{l c c c c c}
\hline
Steps         & DSO    & Inlier point     & Iteration & Iteration \\
              &        & extraction       &    1      &     7     \\
\hline
$AbsRel$      & 0.0868 &  0.0315          & 0.0587    & 0.0602 \\
Ave. $\#$ of points per image &  1485 &  10362   & 125996    & 159237 \\
\hline
\end{tabular}}
\label{ablation_evaluation}
\vspace{-15pt}
\end{table}

\begin{figure}
\vspace{5pt}
\centering
  \includegraphics[width=\linewidth]{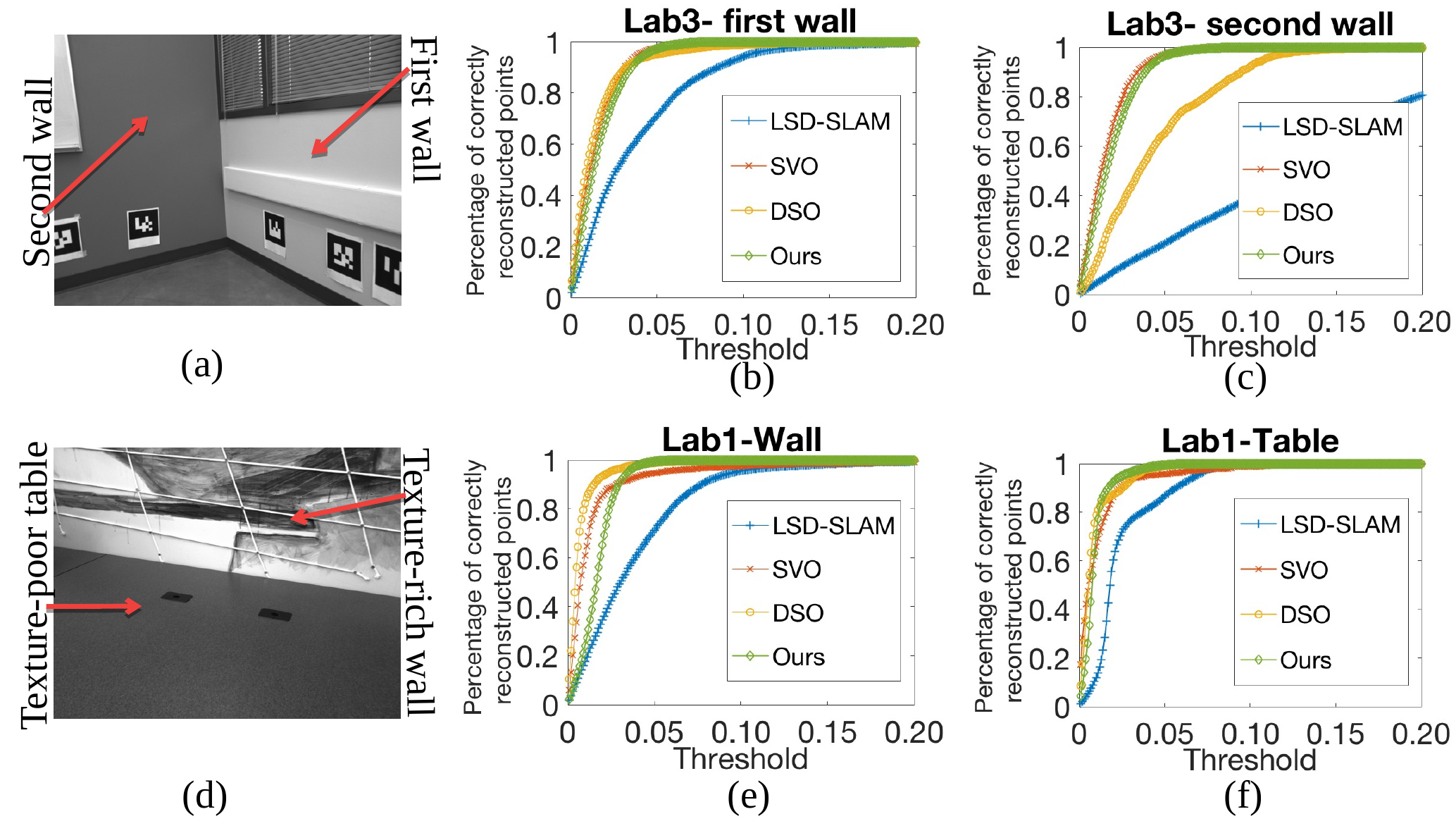}  
\caption{Comparison of results between the proposed method and the visual SLAM methods in terms of the percentage of correctly reconstructed points on sequence ``Lab3" and ``Lab1"}
\label{fig:first_exp}
\vspace{-15pt}
\end{figure}

To show the capability of the proposed method, we also evaluate our method on real images. Since no polarization dataset is available with real images, we chose scenes with planar surfaces (e.g., walls, tables) to evaluate our method. One sample scene for this experiment is shown in Fig.~\ref{fig:first_exp}(a) with two vertical walls. In this experiment, we first segment the final points of the obtained map into two sets that belong to the two vertical walls and remove the points that belong to the floor. Then for each vertical wall, we fit a plane to the points of that wall. As the performance metric, we measure the number of correctly reconstructed points using our method in different scenarios, and compare it with that obtained from the competing visual SLAM methods. By varying the distance to the plane as a threshold, we compute the percentage of correctly estimated points. It is worth mentioning that the two walls contain large texture-less regions.

Figs.~\ref{fig:first_exp}(b) and (c) demonstrate the accuracy of our method in comparison with the competing methods. Both SVO and DSO are known to reconstruct points from point features in a scene, with a relatively high accuracy. However, they are not able to provide a dense map especially in texture-less or texture-poor regions. LSD-SLAM, on the other hand, provides a denser map but with a lower accuracy in comparison with SVO and DSO. As Fig.~\ref{fig:first_exp}(b) shows, our polarimetric method is able to reconstruct a very dense map with a comparable accuracy to DSO or SVO. 

To show the capability of our polarimetric method, we also evaluate our algorithm in a scene with both texture-rich and texture-poor regions (Fig.~\ref{fig:first_exp}(d)), and the evaluation results are shown in Figs.~\ref{fig:first_exp}(e) and (d). This experiment shows that in some cases our method can reconstruct a dense map with an even higher accuracy (e.g., ``table" from Lab1 sequence) than the competing visual SLAM methods, due to the use of total variation optimization on points with mostly one reflection type. In other words, if scene points have mixed reflection, the obtained polarimetric cues may not be completely accurate. 
In fact, points with one or the other dominant reflection allow our method to generate accurate polarimetric cues. Therefore, the propagation and the estimation steps followed by the use of total variation optimization have less error.


Table~\ref{table1} compares our polarimetric method with the competing methods in terms of the number of estimated points. The results show that the map density of our proposed method is a couple orders of magnitude higher than the other methods, as expected.


\subsection{Map reconstruction} 
\begin{figure}[t]
  \centering
  \vspace{5pt}
  \includegraphics[width=0.99\linewidth, height=1.8cm]{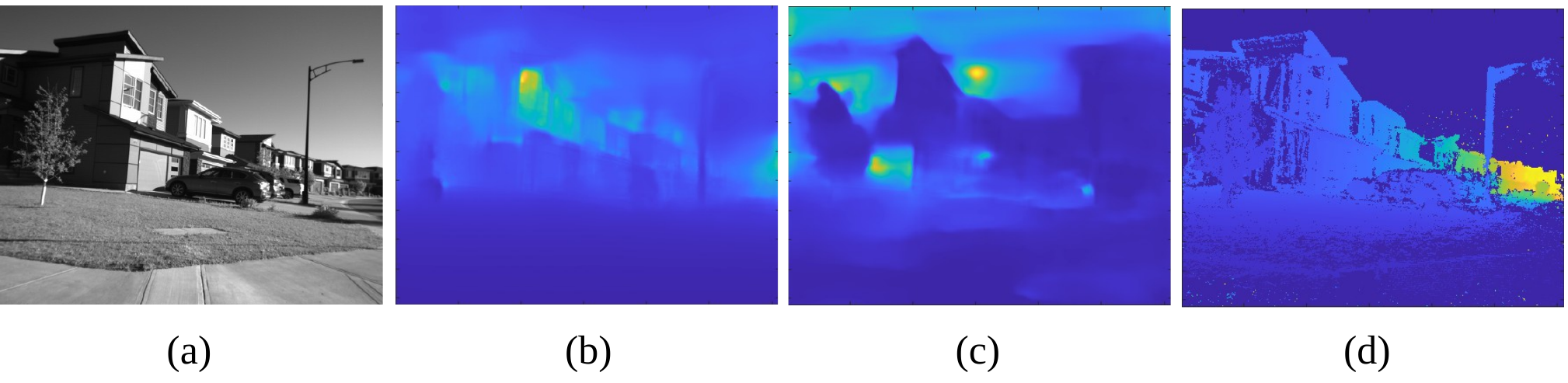}  
  \caption{(a) Input frame, (b)-(d): the obtained depthmap using (b) pre-trained PackNet, (c) fine-tuned PackNet with our data, and (d) our proposed method, respectively.}
  \label{fig:outdoor2}
  \vspace{-15pt}
\end{figure}

In the second set of experiments, we compare our polarimetric method with the state-of-the-art learning based algorithms including ``PackNet"~\cite{PackNet} for outdoor scenes, and ``Im2pcl"~\cite{Im2pcl} and ``DenseNet161"~\cite{densenet} for indoor scenes. Fig.~\ref{fig:outdoor2} shows the qualitative comparison between the proposed method and~\cite{PackNet} in terms of depthmap estimation for a sample frame. As discussed in Section~\ref{related_work}, generalization is a major weakness of learning-based methods. Fig.~\ref{fig:outdoor2}(b) clearly illustrates this issue on PackNet~\cite{PackNet}, a well-known learning-based method. Since PackNet is known for self-supervised depth learning, we also fine-tuned the network with 500 images of our collected dataset; however, the network is still not able to provide meaningful results on out-of-domain data (Fig.~\ref{fig:outdoor2}(c)), and may need more than 500 images. In contrast, our proposed method is able to propagate the depth of sparse points using polarimetric cues in a proper direction to densify the depthmap. 

\begin{figure}[b]
  \centering
  \vspace{-5pt}
  \includegraphics[width=0.99\linewidth, height=4.5cm]{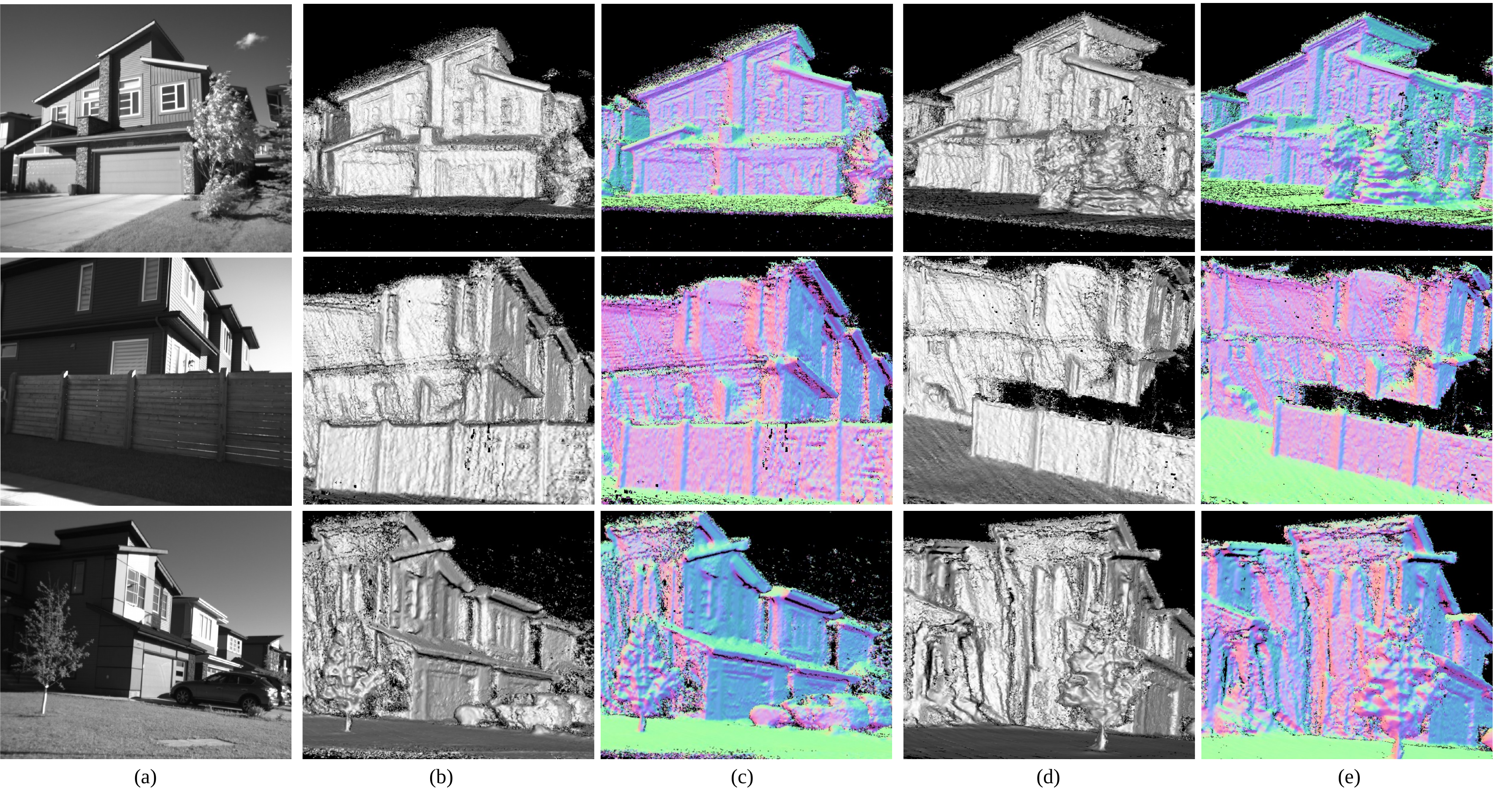}  
  \caption{Dense reconstructed results of our proposed method on three sample scenes. (a) sample image from each scene, (b) and (c) the reconstructed mesh and the corresponding surface normal for better visualization, respectively. (e) and (d) the reconstructed mesh and surface normal from the second view.}
  \label{fig:close_view1}
  \vspace{0pt}
\end{figure}

\begin{table}[b]
\vspace{-5pt}
\caption{Comparison of number of reconstructed points in ``Lab1" and ``Lab3" sequences.} 
\resizebox{\linewidth}{!}
{\begin{tabular}{l c c c r}
\hline
Methods & LSD-SLAM & SVO & DSO & Ours \\
\hline
Lab1 & 192978 & 237 & 494 &  \textbf{370288}\\
Lab3 & 182931 & 186 & 654 &  \textbf{804179}\\
\hline
\end{tabular}}
\label{table1}
\end{table}
Although our proposed method provides less dense results than PackNet, our method is able to produce more accurate and desirable results than PackNet, qualitatively. In addition, some of the holes in each individual depthmap can be generated from subsequent images of the sequence in the fusion step. As an example, Fig.~\ref{fig:close_view1} shows 3D reconstructed results after the fusion step on consecutive depthmaps in three sample scenes. Since the obtained depth from PackNet is not consistent, the fusion algorithm cannot converge to a meaningful result and so, we only show the reconstruction results of our proposed method. Figs.~\ref{fig:close_view1}(a)-(e) illustrate a sample frame from each sequence, 3D reconstructed mesh, and the obtained surface normal of the mesh for better visualization from two views, respectively. These results clearly show that the polarimetric cues with the help of relative depth prior successfully reconstruct the map of outdoor scenes even with texture-poor regions (e.g., garage doors and side walls of buildings).


\begin{figure}[t]
  \centering
  \vspace{5pt}
  \includegraphics[width=\linewidth]{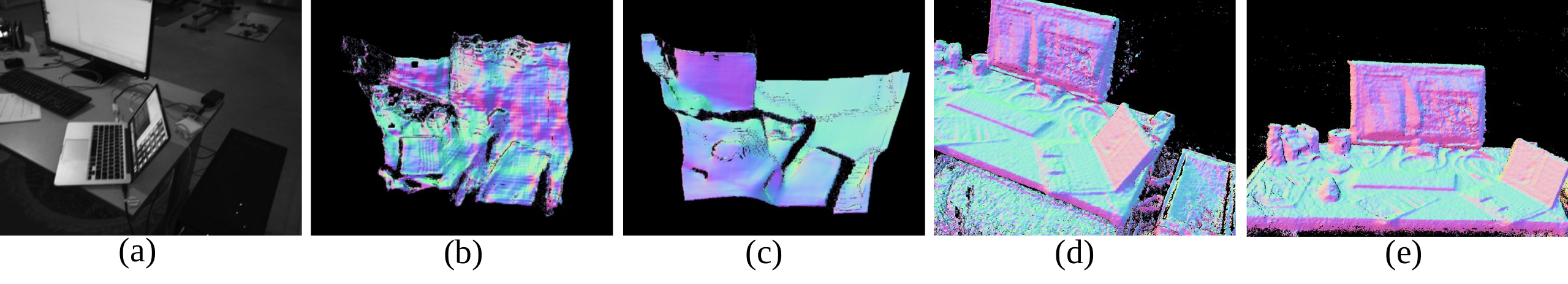}  
  \caption{(a) sample image of an indoor scene. (b)-(e) the reconstructed results of DenseNet161, Im2pcl, and the proposed method at two views.\vspace{-5pt}} 
  \label{fig:indoor1}
  \vspace{-12pt}
\end{figure}

\begin{figure*}[t]
  \centering
  \vspace{-10pt}
  \includegraphics[width=\linewidth]{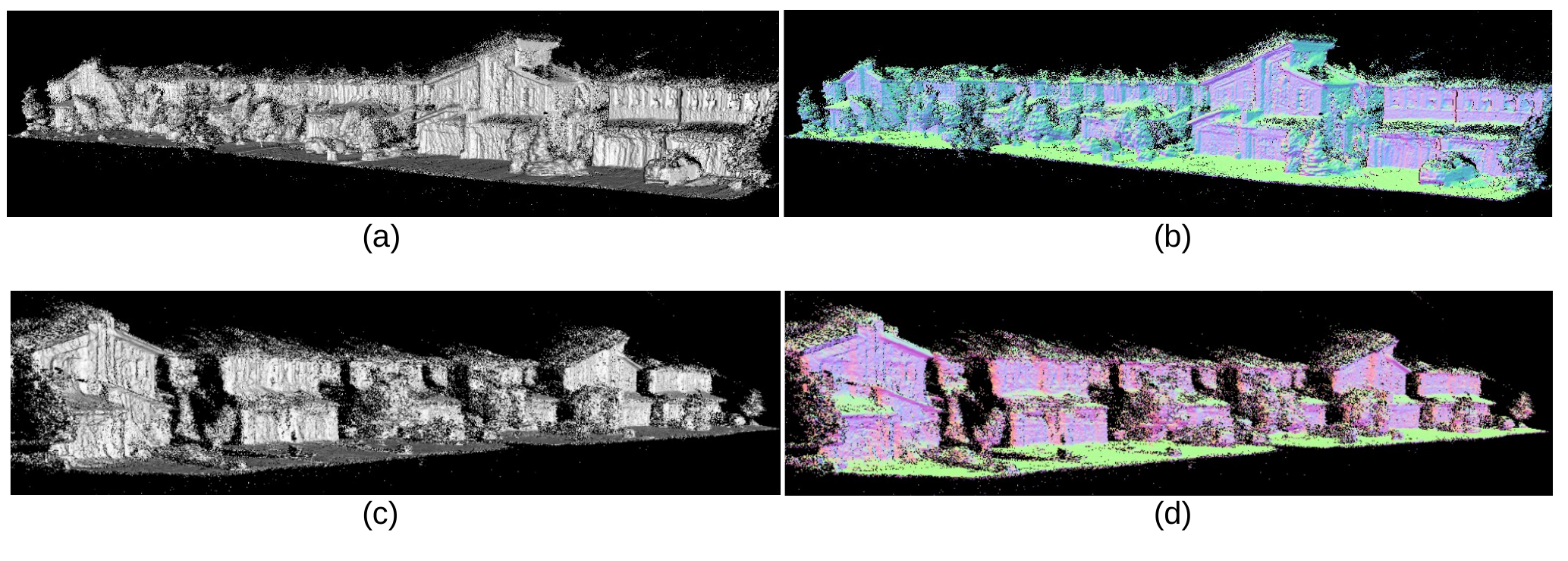}  
  \caption{Dense reconstructed map of sequence ``car1" from two views. (a) and (b): The reconstructed mesh and normals in the original direction. (c) and (d): Second view of the reconstructed map.}
  \label{fig:reconstructed_map3}
\end{figure*}

We also compare our polarimetric method qualitatively with two other deep methods, ``Im2pcl" and ``DenseNet161", on indoor scenes. Fig.~\ref{fig:indoor1} shows the reconstructed map of our method in comparison with the results of these two deep methods. We only used a couple of consecutive frames to reconstruct the map due to the inconsistency between depthmaps of learning based methods. Although the results of Im2pcl are reasonable due to the similarity between the testing and the training scenes, still the results are skewed and inconsistent between frames in map reconstruction, and the fusion step cannot achieve a meaningful map on the whole sequence. 
Finally to show how well the proposed method works on a large-scale sequence, we mounted the polarization camera on a vehicle and captured the polarized image sequence ``car1". Fig.~\ref{fig:reconstructed_map3} shows the reconstructed map of this sequence. The first row of Fig.~\ref{fig:reconstructed_map3} shows the original direction of captured images. The second row of Fig.~\ref{fig:reconstructed_map3} shows reprojection of the reconstructed map from another perspective. The left side walls of houses are not available in the reprojection, since those space points are not visible in the image sequence. Corresponding surface normals in Figs.~\ref{fig:reconstructed_map3}(b) and (d) are shown in pseudo-color for qualitative evaluation of the quality of the reconstructed points. 
 
\subsection{Time complexity}
\label{time}
In this section, we analyze the time complexity of the proposed method. Since our method forms a pipeline system, we exclude the time complexity of DSO and InfiniTAM, which only affect the system delay but not the throughput. In other words, while the proposed method is processing the current keyframe, both DSO and InfiniTAM can run in real time to process the next and the previous keyframes in their separate threads.
\begin{itemize}
\item{Depth initialization: Initial $z$ and $z'$ are computed in parallel, and the time is dominated by COLMAP with a time complexity of $O(nmp^2)$, where $n=8$ is the number of neighbors, $m$ is the number of pixels in each keyframe, and $p=5$ is the patch size. COLMAP is efficiently implemented to run on a GPU and initializes the depthmap in real-time.}

\item{Disambiguation and inlier point extraction: Each step has the time complexity of $O(m)$.}

\item{Iterative process: This step has the time complexity of $O(i(lm + rm))$, where $l$ is the number of pixels we trace in the propagation/estimation step, $r=3$ is the number of iterations in total variation optimization, and $i$ is the number of iterations in the iterative process.}
\end{itemize}
Therefore, the time complexity of the proposed method is $O(nmp^2+i(lm+rm))$ that is about six times faster than~\cite{polar2} with iterative PatchMatch algorithm. Note that the complexity of the second term (i.e., the iterative process) is less than the complexity of COLMAP and it is suitable for GPU implementation, since the propagation/estimation for the sparse points can be done in parallel. Currently the running time of our method is around 1 second per keyframe on synthetic images, and 4 seconds per keyframe for real images, due to use of CPU for the iterative process taking 3.7s (resolutions of synthetic and real images are $640$x$480$ and $1224$x$1024$, respectively). COLMAP runs on a GeForce RTX 2070S GPU, and the iterative process runs on a AMD Razen Threadripper 2950x CPU with 64GB memory. Although the run time is not ideal, it is a considerable improvement over previous offline methods (e.g.,~\cite{polar1}) which can only work through batch processing. For example, the running time of~\cite{polar1} is around 92 seconds per image in a sequence of 32 images, and increases significantly on larger image sequences.

\section{Conclusion}
\label{conclusion}
In this paper we have proposed a novel method based on polarimetric information for dense map reconstruction. Our method exploits polarimetric cues obtained by a polarization camera to improve the reconstruction of texture-poor regions, which is a long-standing difficulty in computer vision and robotics applications. In our method, we first resolve the ambiguities of polarimetric information using a relative depth map, and use those cues in an iterative precess, which includes depth estimation and propagation, two-view consistency check, and total variation optimization for depth smoothing. Our experimental results on both synthetic and real image sequences show that our method outperforms both sparse and semi-dense visual SLAM algorithms and learning-based methods.

In the proposed method, we used a pre-defined refractive index to estimate the zenith angle, which is reasonable for dielectric materials. However, we can estimate a more accurate refractive index using pairs of inlier points between concecutive keyframes and their camera poses~\cite{polar3}. We leave this as future work. Furthermore, non-dielectric materials have a different formulation for estimating the zenith angles, which leads to additional ambiguities in polarization cues. We plan to investigate this as well in the future.








{\small
\bibliographystyle{ieee_fullname}
\bibliography{egbib}
}

\end{document}